\documentclass{article}
\usepackage{iclr2022_workshop,times}

\usepackage{amsmath,amsfonts,bm}

\def\eqref#1{equation~\ref{#1}}

\def\1{\bm{1}}

\DeclareMathAlphabet{\mathsfit}{\encodingdefault}{\sfdefault}{m}{sl}
\SetMathAlphabet{\mathsfit}{bold}{\encodingdefault}{\sfdefault}{bx}{n}

\usepackage{microtype}
\usepackage{graphicx}
\usepackage{booktabs} 
\usepackage{caption}
\usepackage{subcaption}

\usepackage{hyperref}

\usepackage{amsmath}
\usepackage{amssymb}
\usepackage{mathtools}
\usepackage{amsthm}

\usepackage[capitalize,noabbrev]{cleveref}

\theoremstyle{plain}
\newtheorem{theorem}{Theorem}[section]

\theoremstyle{definition}
\newtheorem{definition}[theorem]{Definition}

\theoremstyle{remark}

\usepackage{hyperref}
\usepackage{url}

\title{Evaluating Disentanglement in Generative Models Without Knowledge of Latent Factors}

\author{Chester Holtz \& Gal Mishne \& Alexander Cloninger\\
University of California San Diego \\
San Diego, USA \\
\texttt{\{chholtz, gmishne, acloninger\}@ucsd.edu} \\
}

\iclrfinalcopy 
\begin{document}

\maketitle

\begin{abstract}
Probabilistic generative models provide a flexible and systematic framework for learning the underlying geometry of data. However, model selection in this setting is challenging, particularly when selecting for ill-defined qualities such as disentanglement or interpretability. In this work, we address this gap by introducing a method for ranking generative models based on the training dynamics exhibited during learning. Inspired by recent theoretical characterizations of disentanglement, our method does not require supervision of the underlying latent factors. We evaluate our approach by demonstrating the need for disentanglement metrics which do not require labels\textemdash the underlying generative factors. We additionally demonstrate that our approach correlates with baseline supervised methods for evaluating disentanglement. Finally, we show that our method can be used as an unsupervised indicator for downstream performance on reinforcement learning and fairness-classification problems.
\end{abstract}

\section{Introduction}
Generative models provide accurate models of data, without expensive manual annotation~\cite{bengio09, kingma2014}. 
However, in contrast to classifiers, 
%
%
fully unsupervised model selection within the class of generative models is far from a solved problem~\cite{locatello2019a}. For example, simply computing and comparing likelihoods can be a challenge for some families of recently proposed models~\cite{goodfellow14,li15}. Given two models that exhibit similar loss-values on a held-out dataset, there is no computationally friendly way to determine whether one likelihood is significantly higher than the other. Permutation testing or other generic strategies are often computationally prohibitive and it is unclear if likelihood correlates with desirable qualities of generative models. We are motivated to address this problem.

In this work, we focus on the problem of unsupervised model selection, namely Variational Autoencoders (VAEs), for \emph{disentanglement}~\cite{higgins17}. In this context, model selection without full or partial supervision of the ground truth generative process and/or attribute labels is currently an open problem and existing metrics exhibit high variance, even for models with the same hyperparameters trained on the same datasets~\cite{locatello2019a, locatello2019b}. Since ground truth generative factors are unknown or expensive to provide in most real-world tasks, it is important to develop efficient unsupervised methods.

To address the aforementioned issues, we propose a simple and flexible method for fully unsupervised model selection for VAE-based disentangled representation learning. Our approach is inspired by recent findings that attempt to explain why VAEs disentangle~\cite{rolinek19}. We characterize disentanglement quality by performing pairwise comparisons between the training dynamics exhibited by models during gradient descent. We validate our approach using baselines discussed in \citet{{locatello2019a, locatello2019b}} and demonstrate that the model rankings produced by our approach correlate well with performance on downstream tasks.

\subsection{Contributions}
Our contributions can be summarized as follows:
\begin{enumerate}
    \item We design a novel method for model selection for disentanglement based on the activation dynamics of the decoder observed throughout training.
    \item Notably, our method is fully unsupervised\textemdash our method does not rely on class labels, training supervised models, or ground-truth generative factors.
    \item We evaluate our proposed metric by demonstrating strong correlation with supervised baselines on the dSprites dataset~\cite{dsprites17} and downstream performance on reinforcement learning and classification tasks~\cite{watters2019,locatello2019b}.
\end{enumerate}

\section{Background}
\label{sec:background}
In this section, we review the background notation, the basic variational autoencoder (VAE) framework, and the concept of \emph{disentanglement} in the context of this framework.
\subsection{Variational Autoencoders}
Let $X = \{x_i\}_{i=1}^N$ be a dataset consisting of $N$ i.i.d samples $x_i \in \mathbb{R}^n$ of a random variable $x$. An autoencoder framework is comprised of two mappings: the encoder $\text{Enc}_\phi:\mathbb{R}^n \to Z$, paramaterized by $\phi$, and the decoder $\text{Dec}_\theta :Z \to \mathbb{R}^n$, paramterized by $\theta$. $Z$ is typically termed the \emph{latent space}. In the \emph{variational} autoencoder (VAE) framework, both mappings are taken to be probabalistic and a fixed prior distribution $p(z)$ over $Z$ is assumed. 

The training objective is the marginalized log-likelihood:
\begin{equation}
\sum_{i=1}^n \log p(x_i)
\label{eq:loglikelihood}
\end{equation}
In practice, the parameters of the model; $\phi$ and $\theta$ are jointly trained via gradient descent to minimize a more tractable surrogate: the Evidence Lower Bound (ELBO)
\begin{equation}
\mathbb{E}_{z \sim q(z|x_i)}\log p(x_i | z) - D_{\text{KL}}(q(z | x_i) || p(z))
\label{eq:elbo}
\end{equation}
where the first term corresponds to the reconstruction loss and the second corresponds to the KL divergence between the latent representation $q(z | x_i)$ and the prior distribution $p(z)$, typically chosen to be the standard normal $\mathcal{N}(0,I)$. A significant extension, $\beta$-VAE, proposed by \citet{higgins17} introduces a weight parameter $\beta$ on the KL term:
\begin{equation}
\mathbb{E}_{z \sim q(z|x_i)}\log p(x_i | z) - \beta D_{\text{KL}}(q(z | x_i) || p(z)).
\label{eq:betaelbo}
\end{equation}
The value of $\beta$ is usually chosen to induce certain desireable qualities in the latent representation\textemdash e.g. interpretability or disentanglement~\cite{chen2018, ridgeway2018}. Recent work has also proposed methods for selecting $\beta$ adaptively or according to pre-defined schedules during training~\cite{bowman2016, fu2019}.

\subsection{Supervised Methods for Evaluating Disentanglement}
\begin{figure}[t]
\centering
\begin{subfigure}[]{0.121\linewidth}
\includegraphics[trim={15cm 6cm 0 0},clip,width=\linewidth]{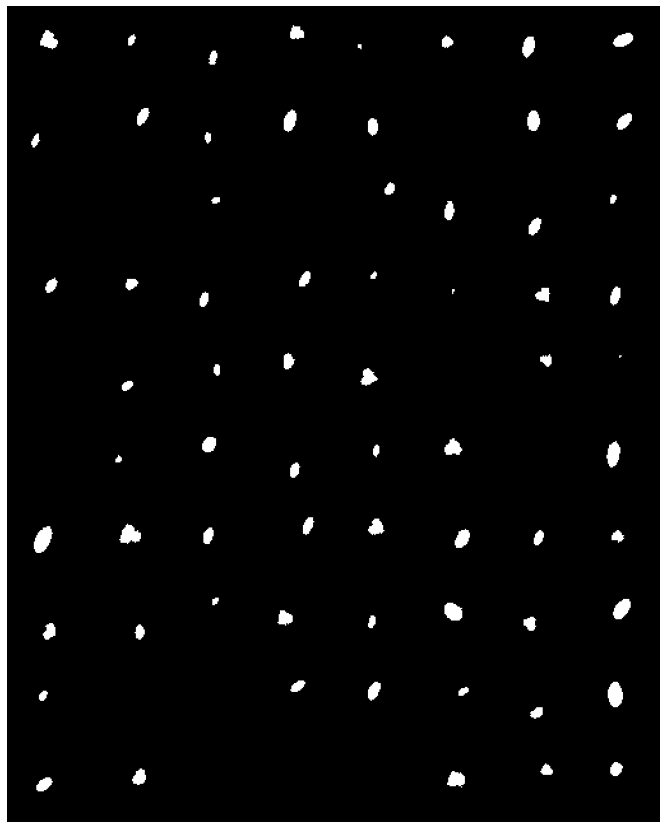}
\subcaption{$\beta = 0.1$}
\label{fig:entangledsprites}
\end{subfigure}
\begin{subfigure}[]{0.124\linewidth}
\includegraphics[width=\linewidth]{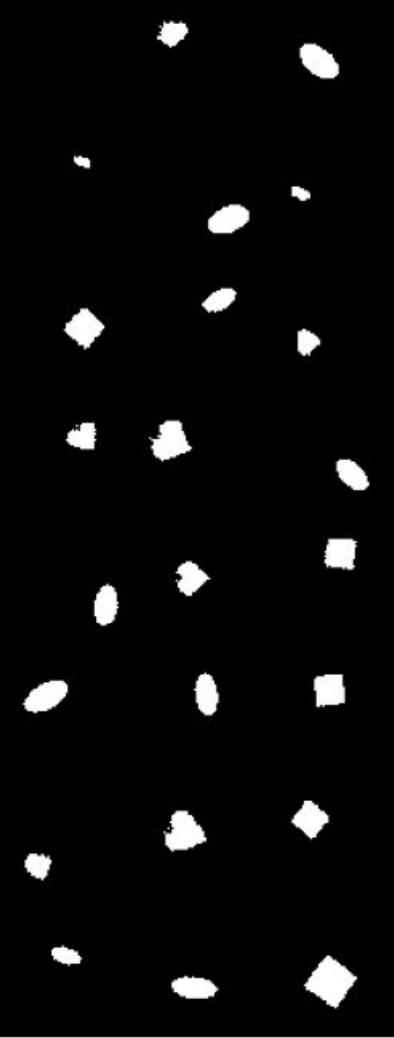}
\subcaption{$\beta = 1.0$}
\label{fig:vsamp4}
\end{subfigure}
\begin{subfigure}[]{0.128\linewidth}
\includegraphics[width=\linewidth]{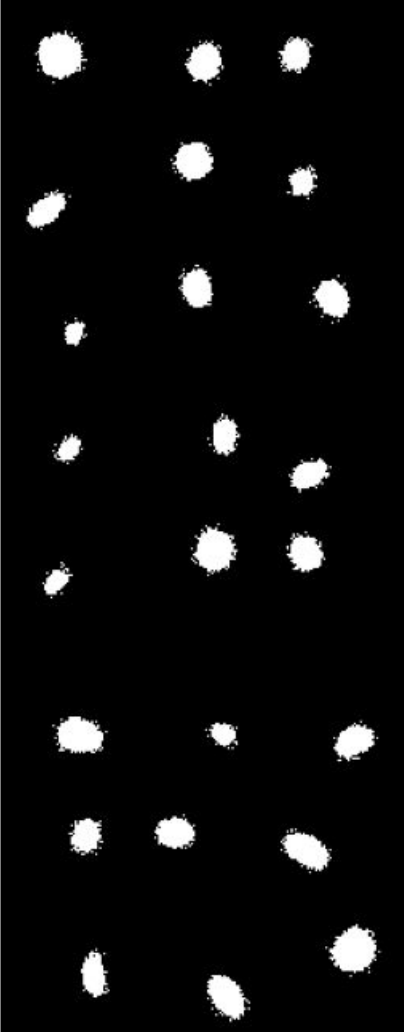}
\subcaption{$\beta = 10$}
\label{fig:vsamp8}
\end{subfigure}
\begin{subfigure}[]{0.12\linewidth}
\includegraphics[width=\linewidth]{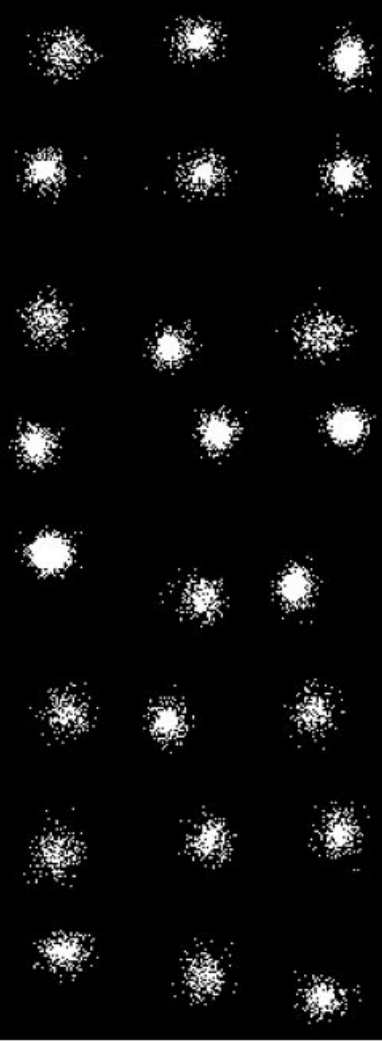}
\subcaption{$\beta = 100$}
\label{fig:vsamp16}
\end{subfigure}
\caption{(\textbf{\subref{fig:entangledsprites}\textemdash\subref{fig:vsamp16}.}) dSprite samples from \emph{entangled} and \emph{disentangled} latent spaces. Note the occurrence of noisy, missing, and unrealistic samples when $\beta$ is set inappropriately ($\beta=0.1, 10, 100$). We propose an unsupervised algorithm to find the appropriate choice of parameters such as $\beta$ to encourage disentanglement.}

\label{fig:dspritesamples}
\end{figure}
There have been recent efforts by the deep learning community towards learning intrinsic generative factors from data, commonly referred to as learning a disentangled representation. While there are few formalizations of disentanglement, an informal description is provided by \citet{bengio09}:
\begin{quotation}\textit{a representation where a change in one dimension corresponds to a change in one factor of variation, while being relatively invariant to changes in other factors.}
\end{quotation}
Recent work has shown that learning disentangled representations facilitate robustness~\cite{Yang_Guo_Wang_Xu_2021}, interpretability~\cite{zhu2021}, and other desireable characteristics~\cite{locatello2019b}. A simple example of the difference between the quality of samples drawn from entangled and disentangled models is provided in Fig.~\ref{fig:dspritesamples}. As a result, evaluating models for their ability to learn disentangled latent spaces has received a large amount of attention in recent years~\cite{locatello2019a}.
%
    
\textbf{$\beta$-VAE \& FactorVAE.} $\beta$-VAE~\cite{higgins17} and FactorVAE~\cite{kim2018} are popular methods for evaluating disentanglement and encouraging learning disentangled representations. As previously mentioned, $\beta$-VAE uses a modified version of the VAE objective with a larger weight ($\beta > 1$) on the KL divergence between the variational posterior and the prior, and has proven to be  effective for encouraging disentangled representations.

In addition to introducing a modification of the ELBO loss, \citet{higgins17} proposed a supervised metric that attempts to quantify disentanglement when the ground truth factors of a data set are given. The metric is the error rate of a linear classifier computed as follows:
\begin{enumerate}
    \item Choose a factor and sample data $x$ with the factor fixed
    \item Obtain their representations (mean of $q(z|x)$)
    \item Take the absolute value of the pairwise differences of these representations.
    \item The mean of these statistics across the pairs are the variables and the index of the fixed factor is the corresponding response
\end{enumerate}
Intuitively, if the learned representations were perfectly disentangled, the dimension of the encoding corresponding to the fixed generative factor would be exactly zero, and the linear classifier would map the index of the zero to the index of the factor. However this metric has several weaknesses:
\begin{enumerate}
    \item The classifier requires labeled generative factors
    \item The metric is sensitive to the classifier's parameters
    \item The coefficients of the classifier may not be sparse
    \item The classifier may give $100\%$ accuracy even when only $k - 1$ factors out of $k$ have been disentangled
\end{enumerate}
In an attempt to resolve the issues resulting from the application of a parametric model, \citet{kim2018} proposed to replace the linear predictor with a nonparametric majority-vote classifier applied to the empirical variances of the latent embeddings. In other words, the classifier predicts the generative factor $k$ corresponding to the latent dimension with the smallest variance. However, although the drawbacks of linear classification are addressed, certain new limitations are introduced: (1.) an assumption of independence between generative factors (2.) necessity of factor labels.

\textbf{Mutual Information Gap. } The Mutual Information Gap (MIG)~\cite{chen2018} metric involves estimating the mutual information between generative factors each latent dimensions. For each factor, \citet{chen2018} consider the pair of latent dimensions with the highest MI scores. It is assumed that in a disentangled representation the difference between these two scores would be large. The MIG score is the average normalized difference between pairs of MI scores. \citet{chen2018} claim that the MIG score is more general compared to the $\beta$-VAE and FactorVAE metrics. However, as with $\beta$-VAE and FactorVAE, the labels of the underlying generative factors are required.

\subsection{Intrinsic Indicators of Disentanglement}
\begin{figure}[t]
\centering
\includegraphics[width=0.6\textwidth]{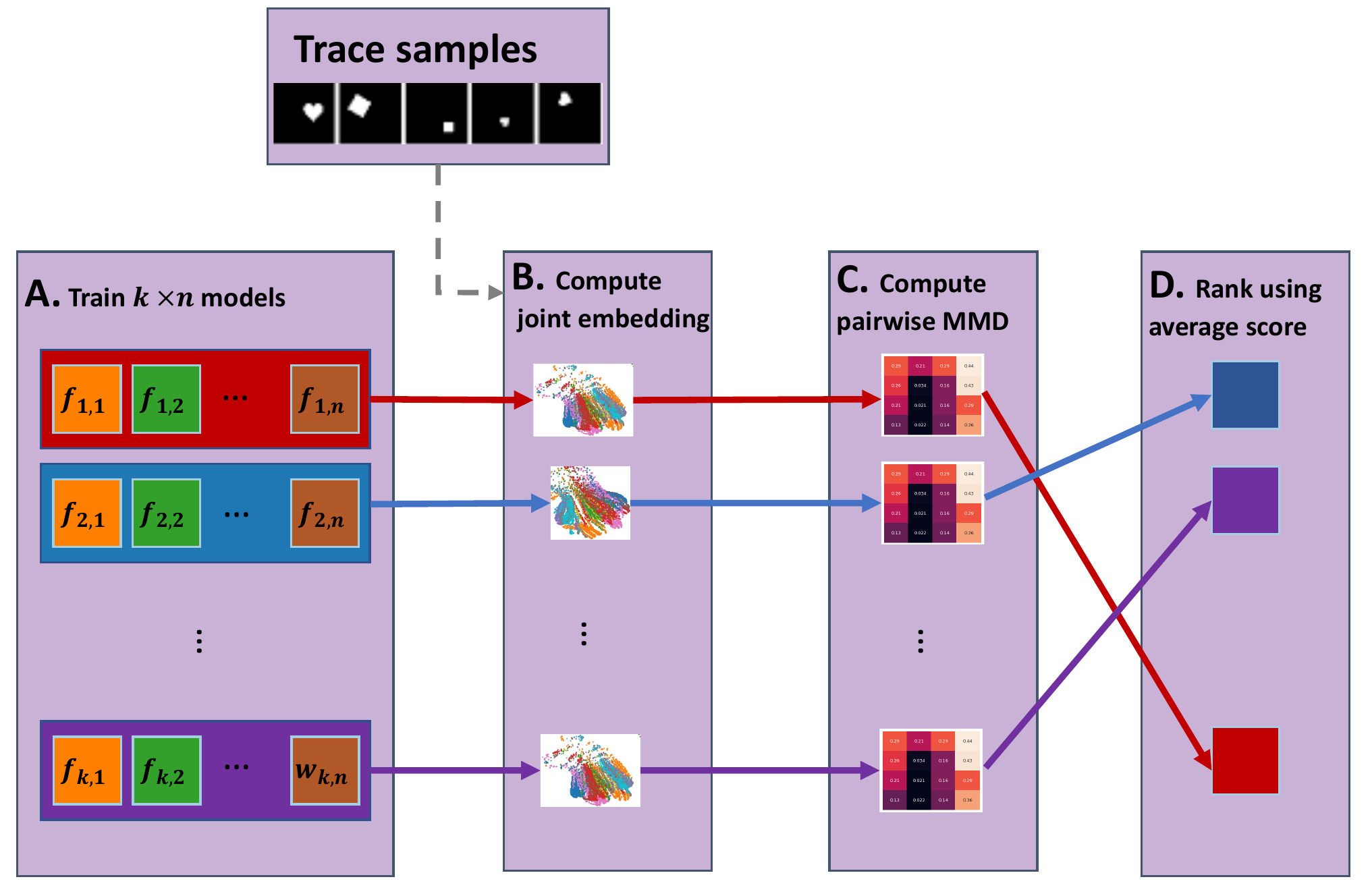}
\caption{Framework. (\textbf{A}) 
Step 1: Model training: $n$ networks are trained from different initializations for each model specification.  (\textbf{B}) Step 2: Networks are jointly embedded according to training dynamics. (\textbf{C}) Step 3: Pairwise MMD scores are computed from the joint embeddings calculated in Step 2. (\textbf{D}) Step 4: The network specifications are sorted using the mean MMD score.}
\label{fig:workflow}
\end{figure}
In this section, we review recent work on identifying fundamental indicators of disentangled models. Recent work \cite{Duan2020Unsupervised,zhou2021evaluating, rotman2022unsupervised, bounliphone2015, khrulkov18a} has explored various unsupervised scoring functions to evaluate and compare generative models for similarity, completeness, and disentanglement from the perspective of the latent space. In particular, \cite{Duan2020Unsupervised} and \cite{zhou2021evaluating} propose unsupervised methods for measuring disentanglement based on computing notions of similarity between generative models. \citet{zhou2021evaluating} utilize persistence homolgy to motivate a topological dissimilarity between latent embedding spaces, while \citet{zhou2021evaluating} propose a statistical test between simpling distributions of latent activations. As far as we are aware, we are the first to exploit the dynamics of activations observed during training.

\citet{he2018lagging} investigate disentanglement by studying the dynamics of the deep VAE ELBO loss observed during gradient descent. Their conclusions suggest that artifacts of poor hyperparameter selection or architecture design, e.g., posterior collapse, are a direct product of a ``mismatch'' between the variational distribution and true posterior. \citet{chechik05, kunin19} showed that suitable regularization allows linear autoencoders to recover principal components up to rotation. \citet{lucas19} explicitly show that linear VAEs with a diagonal covariance structure recover the principal components exactly.

Significantly, \citet{rolinek19} observed that the diagonal covariance used in the variational distribution of VAEs encourages orthogonal representation. They utilize linearaizations of deep networks to rigorously motivate these observations, along with an assumption to handle the presence of posterior collapse. Following up on this work, \citet{kumar20} empirically demonstrate a more general relationship between the variational distribution covariance and the Jacobian of the decoder. In particular, \citet{kumar20} show that a block diagonal covariance structure implies a block structure of the Jacobian of the decoder. 

The proposed method is motivated by these central results.
\begin{enumerate}
    \item Local minima are global
    \item The local linearization of the Jacobian,\\$J_i = \frac{\partial \text{Dec}_\theta (\mu_\phi (x_i))}{\partial \mu_\phi (x_i)}$ is orthogonal
\end{enumerate}
In summary, we design a method to quantify disentaglement according to a novel notion of disagreement between decoder dynamics for multiple instantiations of VAE specifications during learning.

\section{Comparing VAEs via Learning Dynamics}
\label{sec:method}

The two results mentioned above imply that stability of the activation dynamics of the decoder with respect to different initializations may correlate with disentanglement. In this section, we propose a method for computing a similarity score between two decoders according to their activation dynamics. We hypothesize that realizations of a particular specification of a VAE (its architecture and various hyperparameters) which encourages disentangled representation learning will exhibit similar activation dynamics during training, regardless of initialization.

\subsection{Finding a Common Representation}
To compare the dynamics of multiple VAEs, we define a \emph{multislice} kernel defined on the per-epoch activations between a fixed set of samples. 
\begin{figure}[t]
\centering
\begin{subfigure}[b]{0.24\linewidth}
\includegraphics[width=\linewidth]{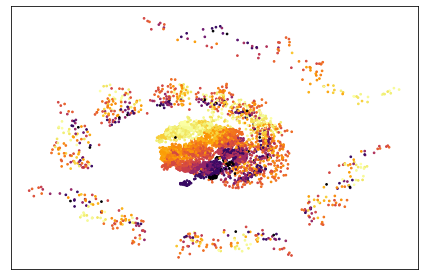}
\subcaption{}
\label{fig:mphate1}
\end{subfigure}
\begin{subfigure}[b]{0.24\linewidth}
\includegraphics[width=\linewidth]{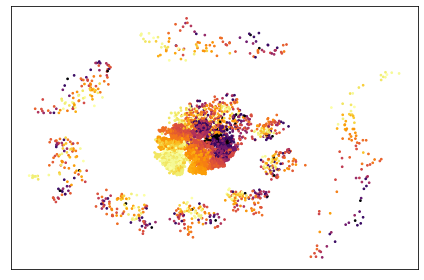}
\subcaption{}
\label{fig:mphate2}
\end{subfigure}
\begin{subfigure}[b]{0.24\linewidth}
\includegraphics[width=\linewidth]{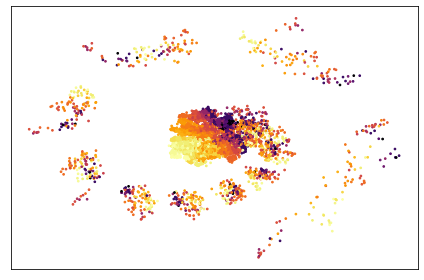}
\subcaption{}
\label{fig:mphate3}
\end{subfigure}
\begin{subfigure}[b]{0.24\linewidth}
\includegraphics[width=\linewidth]{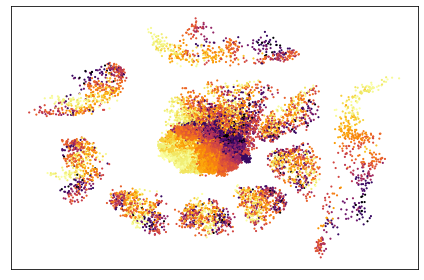}
\subcaption{}
\label{fig:mphate4}
\end{subfigure}
\caption{(\textbf{\subref{fig:mphate1}\textemdash\subref{fig:mphate3}.}) $2$d embeddings of training dynamics of individual network realizations. (\textbf{\subref{fig:mphate4}.}) Joint $2$d embeddings of training dynamics. $x$ and $y$ coordinates are integers between $0$ and $64$. Pairs of coordinates are mapped to single values via row-major order\textemdash i.e. samples are colored according to the value $x+y\cdot64$.}
\label{fig:mphate_alignment}
\end{figure}

\textbf{The Multislice Kernel.} We construct a \emph{Multislice Kernel}~\cite{gigante19} defined over a fixed set of \emph{trace} samples. The entries of the kernel\textemdash i.e. the similarities between input samples\textemdash are computed according to the \emph{intermediate activations} exhibited by the decoder.

Following the notation of \citet{gigante19}, the time trace $\mathbf{T}$ of the decoder is an $n\times m\times p$ tensor encoding the activations at each epoch $\tau \in [1,n]$ of $p$ hidden units $\text{Dec}_{\theta_\tau}$ with respect to each of $m$ trace samples.  
A pair of kernels are constructed using $\mathbf{T}$: $\mathbf{K}_{\text{intraslice}}$, which encodes the affinity between pairs of trace samples indexed by $i$ and $j$ at epoch $\tau$ according to the activation patterns they induce when encoded and passed to $\text{Dec}_{\theta_\tau}$, and $\mathbf{K}_{\text{interslice}}$, which encodes the ``self-affinity'' between a sample $i$ at time $\tau$ and itself at time $\nu$:
\begin{align*}
  &\mathbf{K}_{\text{intraslice}}^{(\tau)}(i,j) =\exp(-||\mathbf{T}(\tau,i) - \mathbf{T}(\tau,j)||^\alpha_2 / \sigma^2_{(\tau,i)})  \\
  &\mathbf{K}_{\text{interslice}}^{(i)}(i,j) = \exp(-||\mathbf{T}(\tau,i) - \mathbf{T}(\nu,i)||^2_2 / \epsilon^2),
\end{align*}
where $\sigma_{\tau,i}$ and $\epsilon$ correspond to intraslice and interslice kernel bandwidth parameters. The multislice kernel matrix $\mathbf{K}$ and its symmetrization $\mathbf{K}'$ are then defined:
\begin{align*}
&\mathbf{K}((\tau,i), (\nu,j)) = 
\begin{cases}
\mathbf{K}_{\text{intraslice}} & \text{if } \tau=\nu\\
\mathbf{K}_{\text{interslice}} & \text{if } i=j\\
0 & \text{otherwise}
\end{cases} \quad \mathbf{K}' = \frac{1}{2} (\mathbf{K} + \mathbf{K}^\top)
\end{align*}
$\mathbf{K}'$ is row-normalized to obtain $\mathbf{P} = \mathbf{D}^{-1}\mathbf{K'}$. 
The row-stochastic matrix $\mathbf{P}$ represent a random walk over the samples across all epochs, where propagating from $(\tau, i)$ to $(\nu, j)$ is conditional on the transition probabilities between epochs $\tau$ and $\nu$~\citep{gigante19}. Powers of the matrix, the \emph{diffusion kernel} $\mathbf{P}^t$, represents running the chain forward $t$ steps. \citet{gigante19} define a distance based on $\mathbf{P}^t$ and a corresponding distance preserving embedding.

%
%

\textbf{Joint embeddings. }It is important to note that \citet{gigante19} originally proposed and applied the Multislice Kernel to characterize and differentiate the behavior of different classifiers by constructing visualizations of the network's \emph{hidden units} based on their activations on a fixed set of training samples. In contrast, we propose to apply the Multislice Kernel to directly compare variational models according to the activation response of the decoder on a fixed set of \emph{trace samples}. In other words, in the work of \citet{gigante19}, the entries of $\mathbf{K'}$ correspond to similarities between \emph{hidden units}, but in our method, the entries of $\mathbf{K'}$ correspond to similarities between \emph{trace samples}. This subtle difference is key and implies a simple and direct method to compare different VAEs with \emph{different} architectures by comparing their associated multi-slice kernels computed on the \emph{same} set of trace samples. We accomplish this by concatenating the rows of diffusion kernels associated with each set of realizations per-specification and computing the left singular vectors of this tall matrix. 
Inspired by \citet{gigante19}, we expect that each individual the kernel encode some average sense of affinity across the data, and by extension that the matrix derived by concatenating these kernels is similarly meaningful. 

In Fig.~\ref{fig:mphate_alignment}, we plot the embeddings associated with a single realizations (initializations) of a given model specification (Fig.~\ref{fig:mphate_alignment}~\subref{fig:mphate1}\textemdash\subref{fig:mphate3}) and the aligned embeddings (left singular vectors of concatenated kernels, Fig.~\ref{fig:mphate_alignment}~\subref{fig:mphate4}). Note that each embedding is a slight perturbation of the others, but sample embeddings in the joint space roughly align according to a generative factor that explains a significant amount of sample variance (coordinate position).

\subsection{Maximum Mean Discrepancy (MMD)}

Recall that we propose to consider the cumulative kernel similarity between independent realizations of a VAE specification as a proxy for the disentanglement. To compute a similarity between joint embeddings, we apply the Maximum Mean Discrepancy (MMD) test statistic~\cite{gretton12a} to the left singular vectors of the matrix formed by concatenating diffusion kernels.

\begin{definition}{Maximum Mean Discrepancy (MMD;~\citet{gretton12a}) }
Let $\mathcal{F}$ be a Reproducing Kernel Hilbert space (RKHS), with continuous feature mapping $\phi(x) \in \mathcal{F}$ from each $x \in \mathcal{X}$, such that the inner product between the features is given by the kernel function $k(x,x'):=  \langle \phi(x), \phi(x') \rangle$. Then the squared population MMD is 
\begin{align*}
\text{MMD}^2 (\mathcal{F}, P_x, P_y) = \mathbb{E}_{x,x'}[k(x,x')] - 2\:\mathbb{E}_{x,y\phantom{'}}[k(x,y\phantom{'})] + 
\mathbb{E}_{y,y'}[k(y,y')].
\end{align*}
\end{definition}
To summarize, distances between distributions are represented as distance between mean embeddings of features characterized by the map $\phi$. 
\subsection{Unsupervised Model Selection for Disentanglement}

As mentioned previously, we are motivated by the observation that networks which disentangle well exhibit ``stable'' learning dynamics under different initializations. We approximate this stability with the similarity between learning dynamics for networks that differ \emph{only} in their initial weights. We characterize the learning dynamics according the principles proposed by \citet{gigante19}

\begin{figure}[t]
\begin{subfigure}[b]{0.24\linewidth}
\includegraphics[width=\linewidth]{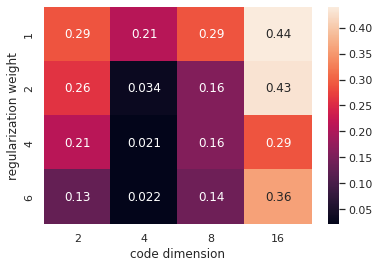}
\subcaption{}
\label{fig:mmd_heatmap}
\end{subfigure}
\begin{subfigure}[b]{0.245\linewidth}
\includegraphics[width=\linewidth]{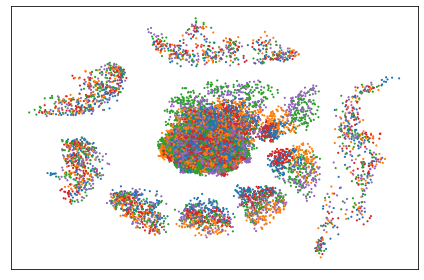}
\subcaption{}
\label{fig:mphate1ct}
\end{subfigure}
\begin{subfigure}[b]{0.245\linewidth}
\includegraphics[width=\linewidth]{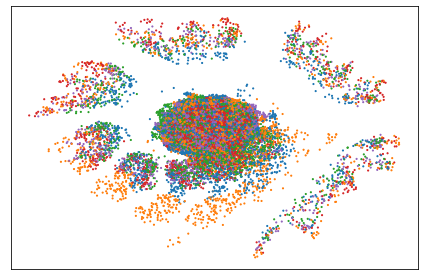}
\subcaption{}
\label{fig:mphate2ct}
\end{subfigure}
\begin{subfigure}[b]{0.245\linewidth}
\includegraphics[width=\linewidth]{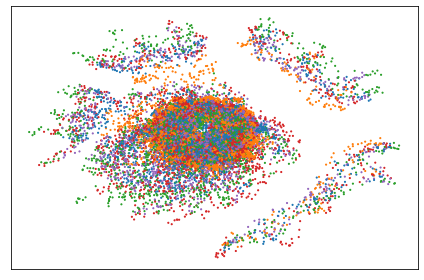}
\subcaption{}
\label{fig:mphate3ct}
\end{subfigure}

\caption{(\textbf{\subref{fig:mmd_heatmap}.}) Average MMD values for different choices of the dimension of $z$ (dimension of the latent space). A lower value denotes more stable learning dynamics. Note that $4$ is the ground truth number of generative factors. (\textbf{\subref{fig:mphate1ct}\textemdash\subref{fig:mphate3ct}.}) Joint $2$-d embeddings of network dynamics as the number of latent dimensions ranges from $4,8,16$. Colors denote weight initializations. }
\label{fig:score_dimension}
\vspace{-0.5cm}
\end{figure}

Our method consists of four steps below and in Fig.~\ref{fig:workflow}.\looseness=-1
\begin{enumerate}
    \item Train $k\times n$ different models ($k$ different ``specifications'': architectures / hyperparameters, $n$ different random ``realizations'' per instance)
    \item Jointly embed each group of $n$ models using the left singular vectors of the concatenated multislice kernels
    \item For each group, calculate the pairwise MMD metric between each of pair of $n$ models
    \item Report the average of the MMD metric over each group as the score for the corresponding realization
\end{enumerate}
An example of the above algorithm applied to a set of VAEs which differ in architecture and regularization weight $\beta$ is provided in Fig.~\ref{fig:score_dimension}~\subref{fig:mmd_heatmap}. Note that the scores are smallest for networks with a latent space whose dimension is equal to the number of generative factors ($4$), and for fixed dimension, the scores generally increase as the regularization weight increases\textemdash agreeing with previous work~\cite{higgins17}. In Fig.~\ref{fig:score_dimension}~\subref{fig:mphate1ct}\textemdash\subref{fig:mphate3ct} we provide the 2-d restriction of our algorithm to a set of networks with fixed $\beta=1$ with latent dimension chosen from $4,8,16$.

\section{Experiments}

\label{sec:experiments}
We evaluate the proposed method on the dSprites dataset~\cite{dsprites17}. This dataset consists of binary images of individual shapes. Each image in the dataset can be fully described by four generative factors: shape ($3$ values), $x$ and $y$ position ($32$ values), size ($6$ values), and rotation ($40$ values). The generative process for this dataset is fully deterministic, resulting in $737,280$ images. We adopt the same convolutional encoder-decoder architecture presented in~\citet{higgins17}. Network instances vary with respect to the dimension of the code and the $\beta$-factor used during training with $\beta$ chosen from the set $[1,2,4,8,16]$ and latent space dimension in $[2, 4, 8, 16, 32]$.

\begin{table}[h]
\caption{Spearman rank correlations between rankings produced by our method and those produced by three supervised methods for increasing number of initializations per model.}
\label{tab:rankcorr}
\centering
\resizebox{0.46\textwidth}{!}{%
\begin{tabular}{|l| l l l|} 
\hline

\hline

\textit{\# random seeds}       & $n=5$ & $n=10$  & $n=25$          \\ 

\hline
$\beta$-VAE    &  $0.54\pm 0.12$   & $0.57\pm 0.06$  & $0.61 \pm 0.02$ \\
FactorVAE  &  $0.56\pm 0.21$   & $0.60\pm 0.09$  & $0.60 \pm 0.04$   \\
MIG  & $0.60\pm0.07$  &  $0.65\pm0.03$   & $0.69\pm0.02$   \\
\hline
\end{tabular}}
\end{table}
\textbf{Correlation with supervised disentanglement metrics. } We first demonstrate that our method produces rankings that are correlated with those produced using \emph{supervised} baseline methods\textemdash methods that exploit supervision of latent factors. 

Although the proposed method does not require supervision, it does necessitate training multiple realizations of each model specification. To make a fair comparison with existing methods, we compute the mean of the supervised disentanglement scores over each set of networks.\looseness=-1

In Table~\ref{tab:rankcorr}, we show that rankings produced using our score correlate positively with rankings produced using the supervised methods. Furthermore, we observe that as the number of networks used to compute the score increases, the correlation and standard deviation improves.
%
\begin{table}[h]
\caption{Parameters of noise regimes.}
\label{tab:noiseregimes}
\begin{center}
\begin{tabular}{lllll}
\hline
Noise Model & $p_{\text{shape}}$ & $p_{\text{scale}}$ & $p_{\text{orient.}}$ & $p_{\text{pos.}}$ \\
\hline 
Noise model 1& $0.05$ & $0.1$ & $0.05$  & $0.05$ \\
Noise model 2 & $0.1$ & $0.2$ & $0.1$  & $0.1$ \\
Noise model 3 & $0.3$ & $0.3$ & $0.3$  & $0.3$ \\
\hline
\end{tabular}
\end{center}
\end{table}

\textbf{A failure mode of supervised methods. } In many real world datasets, the generative factors are unknown or are unreliably labeled. We demonstrate that methods which rely on supervision of the latent factors are brittle to label noise. We propose a new instance of dSprites: \emph{noisy-dSprites}. We compare the robustness of various metrics on the dSprites dataset with labels diluted with different amounts of noise. More concretely, when selecting samples with fixed generative $k$ (i.e. step 1. of the $\beta$-VAE metric), we perturb factor $k$, either uniformly if the factor is discrete (e.g. shape) or according to Gaussian noise with mean $0$ (size, orientation, position). We introduce three \emph{noise regimes} in Table~\ref{tab:noiseregimes}. In table~\ref{tab:noisemodel}, for various disentanglement metrics, we show that as the amount of noise increases, the quality of the metric decays. However, since the proposed method does not require labeled generative factors, it is robust to label noise.
\begin{table}[h]
\caption{Pearson's correlation between disentanglement scores of models evaluated on data with noisy factors and scores of models evaluated using true labels. A larger correlation implies robustness.}
\label{tab:noisemodel}
\centering
\resizebox{0.4\textwidth}{!}{%
\begin{tabular}{|l| l l l|} 
\hline

\hline

noise model       & model 1 & model 2  & model 3          \\ 

\hline
Ours  & 1.0   &  1.0   & 1.0  \\
$\beta$-VAE    & 0.69    &  0.54  & 0.47 \\
FactorVAE  & 0.84    &  0.75  & 0.63    \\
MIG  & 0.86  &  0.76   & 0.67  \\
\hline
\end{tabular}}
\end{table}
\begin{table}[!h]
\caption{Correlations between disentanglement metrics and unfairness score from~\citet{locatello2019b} (smaller is better) and sample efficiency of a reinforcement learning agent on a toy clustering task~\cite{watters2019} (smaller is better).}
\label{tab:downstream}
\centering
\resizebox{0.38\textwidth}{!}{%
\begin{tabular}{|l| l l|} 
\hline

\hline

\textit{\# random seeds}       & unfairness & clustering          \\ 

\hline
Ours    &  $-0.72$   & $-0.59$  \\
$\beta$-VAE    &  $-0.75$   & $-0.51$  \\
FactorVAE  &  $-0.80$   & $-0.56$     \\
MIG  & $-0.66$  &  $-0.48$    \\
\hline
\end{tabular}}
\end{table}

\textbf{Correlation with downstream task performance. }  It has been shown that learning (e.g. classifiers or RL agents) with disentangled representations~\cite{watters2019,locatello2019b} is easier in some sense\textemdash i.e. online decision making can be done more efficiently (with respect to sample complexity) when the state-space is disentangled~\cite{watters2019}. Here, we demonstrate that our method can be used to identify VAEs that are useful for downstream classification tasks where training data efficiency is important. More precisely, we evaluate efficiency as the number of steps needed to achieve 90\% accuracy on a clustering task.\looseness=-1

The agent is provided with a pre-trained encoder trained with $\beta \in \{0, 0.01, 0.1, 1\}$, an exploration policy and a transition model. The goal is to learn a reward predictor to cluster shapes by various generative factors. We use 5 random initialisations of the reward predictor for each possible MONet model, and train them to perform the clustering task detailed in \citet{watters2019}.

We additionally evaluate our method according to it's fairness as defined in~\citet{locatello2019b}:
$$
\text{unfairness}(\hat{y}) = \frac{1}{|S|}\sum_sD_\text{TV}(p(\hat{y}), p(\hat{y} | s = s))\quad \forall y
$$
We adopted a similar setup described in \citet{locatello2019b}. A gradient boosted classifier is trained over $10000$ labelled examples. The fairness score is computed by taking the mean of the fairness scores across all targets and all sensitive variables where the fairness scores are computed by measuring the total variation.

In Table~\ref{tab:downstream}, we see that our method exhibits high Spearman correlation with the fairness score and superior correlation with sample efficiency on the reinforcement learning task.

\section{Conclusion and Future Work}

We have introduced a method for unsupervised model selection for variational disentangled representation learning. We demonstrated that our metric is reliably correlated with three baseline supervised disentanglement metrics and with performance on two downstream tasks. Crucially, our method does not rely on supervision of the ground-truth generative factors and is therefore robust to nonexistent or noisily labeled generative factors. Future work includes exploring more challenging datasets, addressing scalability, and integrating labels and adapting our framework to other contexts by exploring qualities of neural networks correlate well with training dynamic stability.
\newpage

\bibliography{pmlr_template}
\bibliographystyle{iclr2022_workshop}

\end{document}